\documentclass{article} %
\usepackage{iclr2019_conference, times}

\usepackage{amsmath,amsfonts,bm}

\def\Figref#1{Figure~\ref{#1}}

\def\Tabref#1{Table~\ref{#1}}

\def\secref#1{section~\ref{#1}}
\def\Secref#1{Section~\ref{#1}}
\def\twoSecrefs#1#2{Sections \ref{#1} and \ref{#2}}
\def\eqref#1{equation~\ref{#1}}
\def\Eqref#1{Equation~\ref{#1}}

\def\1{\bm{1}}

\def\va{{\bm{a}}}

\def\ve{{\bm{e}}}

\def\vh{{\bm{h}}}

\def\vr{{\bm{r}}}

\DeclareMathAlphabet{\mathsfit}{\encodingdefault}{\sfdefault}{m}{sl}
\SetMathAlphabet{\mathsfit}{bold}{\encodingdefault}{\sfdefault}{bx}{n}

\usepackage{color}
\usepackage{url}
\usepackage{avm}
\usepackage{booktabs}
\usepackage{multirow}
\usepackage{graphicx}

\usepackage[colorinlistoftodos]{todonotes}
\usepackage[colorlinks=true, citecolor=blue]{hyperref}

\avmfont{\sc}
\avmoptions{active,center}
\avmvalfont{\rm}
\avmsortfont{\scriptsize\it}

\title{IncSQL: Training Incremental Text-to-SQL Parsers with Non-Deterministic Oracles} %

\author{Tianze Shi \thanks{Work done during an internship at Microsoft Research.} \\
Cornell University\\
\texttt{tianze@cs.cornell.edu} \\
\And
Kedar Tatwawadi \footnotemark[1]\\
Stanford University\\
\texttt{kedart@stanford.edu} \\
\And
Kaushik Chakrabarti, Yi Mao, Oleksandr Polozov, Weizhu Chen\\
Microsoft Research and Microsoft Business Application Group\\
\texttt{\{kaushik,maoyi,polozov,wzchen\}@microsoft.com} \\
}

\newcommand{\emptyslot}[0]{\epsilon}
\newcommand{\action}[1]{{\small \textsf{#1}}}

\newcommand{\stringliteral}[1]{\text{\rmfamily\begin{normalsize}``\texttt{#1}''\end{normalsize}}}

\iclrfinalcopy %
\begin{document}

\maketitle

\begin{abstract}

We present a sequence-to-action parsing approach for the natural language to SQL task that incrementally fills the slots of a SQL query with feasible actions from a pre-defined inventory. To account for the fact that typically there are multiple correct SQL queries with the same or very similar semantics, we draw inspiration from syntactic parsing techniques and propose to train our sequence-to-action models with non-deterministic oracles. We evaluate our models on the WikiSQL dataset and achieve an execution accuracy of $83.7\%$ on the test set, a $2.1\%$ absolute improvement over the models trained with traditional static oracles assuming a single correct target SQL query. When further combined with the execution-guided decoding strategy, our model sets a new state-of-the-art performance at an execution accuracy of $87.1\%$. 
\end{abstract}

\section{Introduction}
\label{sec:intro}

Many mission-critical applications in health care, financial markets, and business process management store their information in relational databases~\citep{hillestad2005can,ngai2009application,levine2001new}. Users access that information using a query language such as SQL. Although expressive and powerful, SQL is difficult to master for non-technical users. Even for an expert, writing SQL queries can be challenging as it requires knowing the exact schema of the database and the roles of various entities in the query. Hence, a long-standing goal has been to allow users to interact with the database through natural language~\citep{androutsopoulos1995natural,popescu2003towards}. 

The key to achieving this goal is understanding the semantics of the natural language statements and mapping them to the intended SQL. This problem, also known as NL2SQL, was previously understudied largely due to the availability of annotation. Without paired natural language statement and SQL query, a weak supervision approach may be adopted which reduces supervision from annotated SQL queries to answers~\citep{Liang2011acl}. This is a more difficult learning problem. Therefore only with recent release of a number of large-scale annotated NL2SQL datasets~\citep{zhongSeq2SQL2017, data-sql-advising}, we start to see a surge of interest in solving this problem.

Existing NL2SQL approaches largely fall into two categories: sequence-to-sequence style neural ``machine translation '' systems \citep{zhongSeq2SQL2017,dong2018acl} and sets of modularized models with each predicting a specific part of the SQL queries \citep{xu2017sqlnet,yu2018naacl}.
The former class suffer from the requirement of labeling a single ground truth query while multiple semantically equivalent queries exist for each intent.
For example, as noticed by \citet{zhongSeq2SQL2017}, the ordering of filtering conditions in a query does not affect execution but affects generation. %
To account for this, techniques such as reinforcement learning have been used on top of those sequence-to-sequence models.
The second class of models employ a sequence-to-set approach: they first predict table columns present in the query and then independently predict the rest for each column.
This avoids the ordering issue, but makes it harder to leverage inter-dependencies among conditions. 

In this work, we develop a sequence-to-action parsing approach (\Secref{sec:model}) for the NL2SQL problem. It incrementally fills the slots of a SQL query with actions from an inventory designed for this task. 
Taking inspiration from training oracles in incremental syntactic parsing \citep{goldberg13tacl}, we further propose to use non-deterministic oracles (\Secref{sec:oracle}) for training the incremental parsers. 
These oracles permit multiple correct action continuations from a partial parse, thus are able to account for the logical form variations. 
Our model combines the advantage of a sequence-to-sequence model that captures inter-dependencies within sequence of predictions and a modularized model that avoids any standarized linearization of the logical forms. 
We evaluate our models on the WikiSQL dataset and observe a performance improvement of $2.1\%$
when comparing non-deterministic oracles with traditional static oracles.
We further combine our approach and the execution-guided decoding strategy \citep{2018executionguided} and achieve a new state-of-the-art performance with $87.1\%$ test execution accuracy. Experiments on a filtered ATIS dataset in addition confirm that our models can be applied to other NL2SQL datasets.

\section{Task Definition}
\label{sec:task}

\begin{figure}[t]
    \centering
    \includegraphics[width=\textwidth]{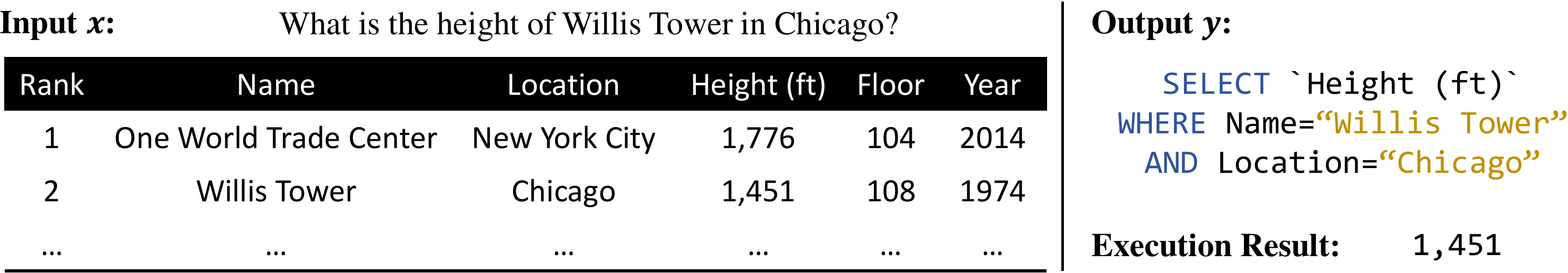}
    \caption{Our running example. The input is a natural language question and a table schema, and the output is an executable SQL query. Table contents are shown here, but unknown to our models. }
    \label{fig:example}
\end{figure}

Given an input natural language question, our goal is to generate its corresponding SQL query.
In the following and throughout the paper, we use the WikiSQL dataset~\citep{zhongSeq2SQL2017} as our motivating example.
However,
it should be noted that our approach is generally applicable to other NL2SQL data, with proper choice of an action inventory and redesign of parser states.

WikiSQL dataset consists of 80,654 pairs of questions and SQL queries distributed across 24,241 tables from Wikipedia. %
Along with the natural language question, the input also contains a single table schema (i.e., table column names).
Each table is present in only one split (train, dev, or test), which demands the models to generalize to unseen tables. \Figref{fig:example} shows an example.

The SQL structure of the WikiSQL dataset queries is restricted and always follows the template \texttt{SELECT agg selcol WHERE col op val (AND col op val)$^*$}. Here, \texttt{selcol} is a single table column and \texttt{agg} is an aggregator (e.g., \texttt{COUNT}, \texttt{SUM}, empty). 
The \texttt{WHERE} segment is a sequence of conjunctive filtering conditions. 
Each \texttt{op} is a filtering operator (e.g., $=$) and the filtering value \texttt{val} is mentioned in the question.
Although the dataset comes with a ``standard'' linear ordering of the conditions, 
the order is actually irrelevant given the semantics of \texttt{AND}. 

Throughout the paper we denote the input to the parser as $x$. 
It consists of a natural language question $w$ with tokens $w_i$ and a single table schema $c$ with column names $c_j$. 
A column name $c_j$ can have one or more tokens. 
The parser needs to generate an executable SQL query $y$ as its output.

\section{Our Model: An Incremental Parser}
\label{sec:model}

Given an input $x$, the generation of a structured output $y$ is broken down into a sequence of parsing decisions. 
The parser starts from an initial state and incrementally takes actions according to a learned policy. 
Each action advances the parser from one state to another, until it reaches one of the terminal states, where we may extract a complete logical form $y$. 
We take a probabilistic approach to model the policy.
It predicts a probability distribution over the valid set of subsequent actions given the input $x$ and the running decoding history. The goal of training such an incremental semantic parser is then to optimize this parameterized policy. 

Formally, we let $P_\theta(y|x) = P_\theta(\va|x)$, where $\theta$ is model parameters. Execution of the action sequence $\va = \{a_1,a_2,\ldots,a_k\}$ leads the parser from the initial state to a terminal state that contains the parsing result $y$. 
Here we assume that each $y$ has only one corresponding action sequence $\va$, an assumption that we will revisit in \Secref{sec:oracle}. 
The probability of action sequence is further factored as the product of incremental decision probabilities: $P_\theta(\va|x)=\prod_{i=1}^{k}P_\theta(a_i|x,a_{<i})$, where $|\va|=k$. 
During inference, instead of attempting to enumerate over the entire output space and find the highest scoring $\va^*=\arg\max_{\va}P_\theta(\va|x)$, our decoder takes a greedy approach: 
at each intermediate step, it picks the highest scoring action according to the policy: $a^*_i=\arg\max_{a_i}P_\theta(a_i|x,a^*_{<i})$.

In the following subsections, we define the parser states and the inventory of actions, followed by a description of our encoder-decoder neural-network model architecture. 

\subsection{Sequence to Actions}
\label{sec:action}

\begin{table}[t]
    \centering
    \small
    \begin{tabular}{lll}
        \toprule
        \textbf{Action} & \textbf{Resulting state after taking the action at state $p$} & \textbf{Parameter representation}  \\
        \midrule
         \action{AGG}(agg) &
         $p$[\textsc{agg}$\mapsto$agg] &
         -- %
         \\
         \action{SELCOL}($c_i$) &
         $p$[\textsc{selcol}$\mapsto c_i$] &
         $\vr^C_i$
         \\
         \action{CONDCOL}($c_i$) &
         $p$[\textsc{cond}$\mapsto p.$\textsc{cond}$||$\begin{avm}
         [col & $c_i$\\op & $\emptyslot$\\val & $\emptyslot$]
         \end{avm}] &
         $\vr^C_i$
         \\
         \action{CONDOP}(op) &
         $p$[\textsc{cond}$_{-1}\mapsto p.$\textsc{cond}$_{-1}$[\textsc{op}$\mapsto$op]] &
         -- %
         \\
         \action{CONDVAL}($w_{i:j}$) &
         $p$[\textsc{cond}$_{-1}\mapsto p.$\textsc{cond}$_{-1}$[\textsc{val}$\mapsto w_{i:j}$]] &
         $\vr^W_i$ and $\vr^W_j$
         \\
         \action{END} &
         $p$ (as a terminal state) &
         --
         \\
        \bottomrule
    \end{tabular}
    \caption{The inventory of actions for our incremental text-to-SQL semantic parser. The use of parameter representations in the decoder is detailed in \Secref{sec:decoder}.
    $p$[\textsc{agg}$\mapsto$agg] denotes a new parser state identical to $p$, except that the feature value for \textsc{agg} is agg in the new state. 
    Finally, $||$~denotes list concatenation, and \textsc{cond}$_{-1}$ refers to the last element in the list.}
    \label{tab:action}
\end{table}

We first look at a structured representation of a full parse corresponding to the example in \Figref{fig:example}:
\begin{center}
\small
\begin{avm}
[agg    & NONE   \\
selcol    & Height \(ft\)     \\
cond      & <[{}col & Name\\op & =\\val & ``Willis Tower''],
            [{}col & Location\\op & =\\val & ``Chicago'']>
]
\end{avm}.
\end{center}

An intermediate parser state is thus defined as a partial representation, with some feature values not filled in yet, denoted as $\emptyslot$. 
The initial parser state $p_0$ has only empty list and values
\begin{avm}
[agg    & $\emptyslot$   \\
selcol    & $\emptyslot$      \\
cond      & <>
]
\end{avm}.

Next, we define our inventory of actions. Each action has the effect of advancing the parser state from $p$ to $p'$. 
We let $p=$ \begin{avm}
[agg    & agg   \\
selcol    & selcol      \\
cond      & cond
]
\end{avm},
and describe $p'$ for each action in \Tabref{tab:action}. 
The action \action{CONDVAL} selects a span of text $w_{i:j}$ from the input question $w$. 
In practice, this leads to a large number of actions, quadratic in the length of the input question, so we break down \action{CONDVAL} into two consecutive actions, one selecting the starting position $w_i$ and the other selecting the ending position $w_j$ for the span. 
At the end of the action sequence, we append a special action \action{END} that terminates the parsing process and brings the parser into a terminal state. As an example, the query in \Figref{fig:example} translates to an action sequence of \{ \action{AGG}(NONE), \action{SELCOL}($c_3$), \action{CONDCOL}($c_1$), \action{CONDOP}($=$), \action{CONDVAL}($w_{5:6}$), \action{CONDCOL}($c_2$), \action{CONDOP}($=$), \action{CONDVAL}($w_{8:8}$)\}.

The above definitions assume all the valid sequences to have the form of \action{AGG} \action{SELCOL} (\action{CONDCOL} \action{CONDOP} \action{CONDVAL})$^*$ \action{END}. 
This guarantees that we can extract a complete logical form from each terminal state.
For other data with different SQL structure, a redesign of action inventory and parser states is required.

\subsection{Decoder}
\label{sec:decoder}

\begin{figure}[t]
    \centering
    \includegraphics[width=\textwidth]{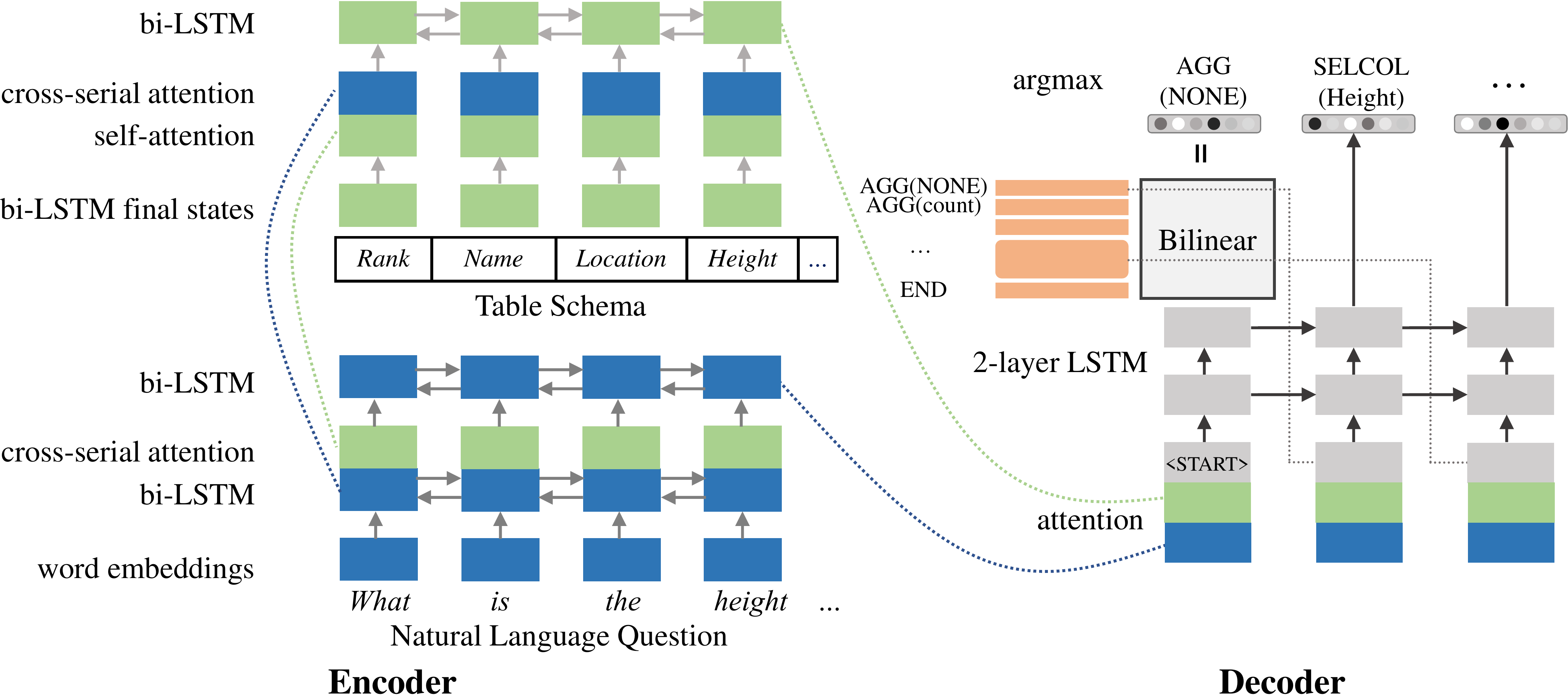}
    \caption{Our main model architecture depicted using our running example. The decoder (right) and the encoder (left) are described in \twoSecrefs{sec:decoder}{sec:encoder}, respectively.
    }
    \label{fig:arch}
\end{figure}

We first assume that we have some context-sensitive representations $\vr^W_i$, $\vr^C_j$ for each word $w_i$ and each column header $c_j$, respectively, and detail our design of the decoder here. 
The encoder for obtaining the representations $\vr^W_i$, $\vr^C_j$ will be discussed in \Secref{sec:encoder}. 

The main component of our decoder is to model a probability distribution $P_\theta(a|x,a_{<i})$ over potential parser actions $a$ conditioned on input $x$ and past actions $a_{<i}$. It has two main challenges:
(1) there is no fixed set of valid parser actions: it depends on the input and the current parser state;
(2) the parser decision is context-dependent: it relies on the decoding history and the information embedded in the input question and column headers. 

We adopt an LSTM-based decoder framework and address the first challenge through individual scoring of actions. 
The model scores each candidate action $a$ as $s_a$ and uses a softmax function to normalize the scores into a probability distribution. 
At time step $i$, we denote the current decoder hidden state as $\vh^{\text{DEC}}_i$ and model the score of $a$ in the form of a bilinear function: $s_a=(\vh^{\text{DEC}}_i)^T U^{A} \vr^{A}_a$,
where $\vr^{A}_a$ is a vector representation of the action $a$
and is modeled as the concatenation of the action embedding and the parameter representation.
The form of the latter is given in \Tabref{tab:action}.

The dependencies between the parser decisions and the input question and column headers are captured through a dot-product attention mechanism \citep{luong2015EMNLP}. 
The input to the first layer of our decoder LSTM at time step $i+1$ is a concatenation of the output action representation $\vr^{A}_{a_i}$ from previous time step $i$, a question attention vector $\ve_i^W$, and a column header attention vector $\ve_i^C$. 
Vector $\ve_i^C=\sum_{j}{\alpha_{i,j}}\vr^C_j$,
where $\alpha_{i,j}\propto \vh^{\text{DEC}}_i \cdot \vr^C_j$. 
Vector $\ve_i^W$ is defined similarly. 

\subsection{Encoder}
\label{sec:encoder}

Now we return to the context-sensitive representations $\vr^W_i$ and $\vr^C_j$. 
Ideally, these representations should be both intra-context-sensitive, i.e. aware of information within the sequences, and inter-sequence-dependent, i.e. utilizing knowledge about the other sequences. 
These intuitions are reflected in our model design as intra-sequence LSTMs, self-attention and cross-serial attention. 

Our model architecture is illustrated in \Figref{fig:arch}. Each word $w_i$ is first mapped to its embedding, and then fed into a bi-directional LSTM (bi-LSTM) that associates each position with a hidden state $\vh^W_i$. For column headers, since 
each column name can have multiple words, we apply word embedding lookup and bi-LSTM for each column name, and use the final hidden state from the bi-LSTM as the initial representation for the column name. 
Next, we apply self-attention \citep{vaswani2017}
to contextualize this initial representation into $\vh^C_j$. 
After obtaining these intra-context-sensitive representations $\vh^W_i$ and $\vh^C_j$, we use cross-serial dot-product attention \citep{luong2015EMNLP}
to get a weighted average of $\vh^C_j$ as the context vector for each $w_i$, and vice versa for each $c_j$. 
The two vectors are concatenated and fed into final bi-LSTMs for the natural language question and table column names, respectively. 
The hidden states of these two LSTMs are our desired context-sensitive representations $\vr^W_i$ and $\vr^C_j$. 

\section{Non-Deterministic Oracles}
\label{sec:oracle}

Previously, we assumed that each natural language question has a single corresponding SQL query, and each query has a single underlying correct action sequence.
However, these assumptions do not hold in practice. 
One well-observed example is the ordering of the filtering conditions in the \textsc{where} clause. Reordering of those conditions leads to different action sequences.
Furthermore, we identify another source of ambiguity in \secref{sec:anycol}, where a question can be expressed by different SQL queries with the same execution results. These queries are equivalent from an end-user perspective. 

For both cases, we obtain multiple correct ``reference'' transition sequences for each training instance and there is no single target policy for our model to mimic during training. 
To solve this, we draw inspiration from syntactic parsing and define {\em non-deterministic} oracles~\citep{goldberg13tacl} that allow our parser to explore alternative correct action sequences. 
In contrast, the training mechanism we discussed in \Secref{sec:model} is called {\em static} oracles. 

We denote the oracle as $O$ that returns a set of correct continuation actions $O(x, a_{<t})$ at time step $t$. Taking any action from the set can lead to some desired parse among a potentially large set of correct results. 
The training objective for each instance $L_x$ is defined as:
\begin{equation}
\label{eq:loss}
L_x=\sum_{i=1}^{k}{\,\log \sum_{a\in O(x,a_{<i})} P_\theta(a|x,a_{<i})},
\end{equation}
where $a_{<i}$ denotes the sequence of actions $a_1,\ldots,a_{i-1}$ %
and $a_i=\arg\max_{a\in O(x,a_{<i})}s_a$,
the most confident correct action to take as decided by the parser during training.
When $O$ is a {\em static} oracle, it always contains a single correct action. 
In that scenario, \Eqref{eq:loss} is reduced to a na\"ive cross-entropy loss. 
When $O$ is {\em non-deterministic}, the parser can be exposed to different correct action sequences and it is no longer forced to conform to a single correct action sequence during training.

\subsection{Alleviating the ``Order-Matters'' Issue}
\label{sec:order-matters}

Training a text-to-SQL parser is known to suffer from the so-called ``order-matters'' issue.
The filtering conditions of the SQL queries do not presume any ordering.
However, an incremental parser must linearize queries and thus impose a pre-defined order.
A correct prediction that differs from a golden labeling in its ordering of conditions then may not be properly rewarded.
Prior work has tackled this issue through reinforcement learning~\citep{zhongSeq2SQL2017} and a modularized \emph{sequence-to-set} solution~\citep{xu2017sqlnet}.
The former lowers optimization stability and increases training time, 
while the latter complicates model design to capture inter-dependencies among clauses: information about a predicted filtering condition is useful for predicting the next condition.

We leverage non-deterministic oracles to alleviate the ``order-matters'' issue. Our model combines the advantage of an incremental approach to leverage inter-dependencies among clauses and the modularized approach for higher-quality training signals. Specifically, at intermediate steps for predicting the next filtering condition, we accept all possible continuations, i.e. conditions that have not been predicted yet, regardless of their linearized positions. 
For the example in \Figref{fig:example}, in addition to the transition sequence we gave in \Secref{sec:action}, our non-deterministic oracles also accept \action{CONDCOL}($c_2$) as a correct continuation of the second action. If our model predicts this action first, it will continue predicting the second filtering condition before predicting the first.

\subsection{Execution-Oriented Modeling of Implicit Column Name Mentions}
\label{sec:anycol}

In preliminary experiments, we observed that a major source of parser errors on the development set is incorrect prediction of implicit column names. 
Many natural language queries do not explicitly mention the column name of the filtering conditions. 
For example, the question in \Figref{fig:example} does not mention the column name ``Name''.
Similarly, a typical question like ``What is the area of Canada?'' does not mention the word ``country''.
For human, such implicit references make natural language queries succinct, and the missing information can be easily inferred from context. But for a machine learning model, they pose a huge challenge.

We leverage the non-deterministic oracles to learn the aforementioned implicit column name mentions by accepting the prediction of a special column name, \texttt{ANYCOL}. 
During execution, we expand such predictions into disjunction of filtering conditions applied to all columns, simulating the intuition why a human can easily locate a column name without hearing it from the query. 
For the example in \Figref{fig:example}, in addition to the action \action{CONDCOL}($c_1$), we also allow an alternative prediction \action{CONDCOL}(\texttt{ANYCOL}). 
When the latter appears in the query (e.g. \texttt{ANYCOL=\stringliteral{Willis Tower}}), we expand it into a disjunctive clause \texttt{(Rank=\stringliteral{Willis Tower} OR Name=\stringliteral{Willis Tower} OR \dots)}. 
With our non-deterministic oracles, when column names can be unambiguously resolved using the filtering values, 
we accept both \texttt{ANYCOL} and the column name as correct actions during training,
allowing our models to predict whichever is easier to learn.

\section{Experiments}
\label{sec:experiments}

\subsection{Dataset and Evaluation}
In our experiments, we use the default train/dev/test split of the WikiSQL dataset. %
We evaluate our models trained with both the static oracles and the non-deterministic oracles on the dev and test split. 
We report both logical form accuracy (i.e., exact match of SQL queries) 
and execution accuracy (i.e., the ratio of predicted SQL queries that result in the same answer after execution). 
The execution accuracy is the metric that we aim to optimize.

\subsection{Implementation Details}
We largely follow the preprocessing steps in prior work of \citet{dong2018acl}. 
Before the embedding layer, only the tokens which appear at least twice in the training data are retained in the vocabulary, the rest are assigned a special ``UNK'' token. 
We use the pre-trained GloVe embeddings \citep{glove}, and allow them to be fine-tuned during training. Embeddings of size $16$ are used for the actions. 
We further use the type embeddings for the natural language queries and column names following \citet{yu2018naacl}: 
for each word $w_i$, we have a discrete feature indicating whether it appears in the column names, and vice versa for $c_j$. 
These features are embedded into $4$-dimensional vectors and are concatenated with word embeddings before being fed into the bi-LSTMs. 
The encoding bi-LSTMs have a single hidden layer with size $256$ ($128$ for each direction). The decoder LSTM has two hidden layers each of size $256$. 
All the attention connections adopt the dot-product form as described in \Secref{sec:decoder}. 

For the training, we use a batch size of $64$ with a dropout rate of $0.3$ to help with the regularization. 
We use Adam optimizer \citep{kingma2014adam} with the default initial learning rate of $0.001$ for the parameter update. 
Gradients are clipped at $5.0$ to increase stability in training. 

\subsection{Results}

The main results are presented in \Tabref{result-table}. 
Our model trained with static oracles achieves comparable results with the current state-of-the-art Coarse2Fine \citep{dong2018acl} and MQAN \citep{McCann2018decaNLP} models.
On top of this strong model, using non-deterministic oracles during training leads to a large improvement of $2.1\%$ in terms of execution accuracy. 
The significant drop in the logical form accuracy is expected, as it is mainly due to the use of \texttt{ANYCOL} option for the column choice:
the resulting SQL query may not match the original annotation. 

We further separate the contribution of ``order-matters'' and \texttt{ANYCOL} for the non-deterministic oracles. When our non-deterministic oracles only address the ``order-matters'' issue as described in \Secref{sec:order-matters},
the model performance stays roughly the same compared with the static-oracle model.
We hypothesize that it is because the ordering variation presented in different training instances is already rich enough for a vanilla sequence-to-action model to learn well.
Adding \texttt{ANYCOL} to the oracle better captures the implicit column name mentions and has a significant impact on the performance, increasing the execution accuracy from $81.8\%$ to $83.7\%$.

Our incremental parser uses a greedy strategy for decoding, i.e. picking the highest scoring action predicted by the policy. A natural extension is to expand the search space using beam search decoding. 
We further incorporate the execution-guided strategy~\citep{2018executionguided} along with beam search. %
The execution-guided decoder avoids generating queries with semantic errors, i.e. runtime errors and empty results. 
The key insight is that a partially generated output can already be executed using the SQL engine against the database, and the execution results can be used to guide the decoding. 
The decoder maintains a state for the partial output, which consists of the aggregation operator, selection column and the completed filtering conditions until that stage in decoding. 
After every action, the execution-guided decoder retains the top-$k$ scoring partial SQL queries free of runtime exceptions and empty output. 
At final stage, the query with the highest likelihood is chosen. 
With $k=5$, the execution-guided decoder on top of our previous best-performing model achieves an execution accuracy of $87.1\%$ on the test set, setting a new state of the art.

\begin{table}[t]
\centering
\small
\begin{tabular}{lcccc}
\toprule
\multirow{2}{*}{Model}& \multicolumn{2}{c}{Dev} & \multicolumn{2}{c}{Test} \\ \cmidrule{2-5}
  & Acc$_\text{lf}$    & Acc$_\text{ex}$ & Acc$_\text{lf}$    & Acc$_\text{ex}$ \\
\midrule
Coarse2Fine~\citep{dong2018acl} & $72.5$ & $79.0$ & $71.7$ & $78.5$\\
MQAN~\citep{McCann2018decaNLP} & $\mathbf{76.1}$ & $82.0$ & $75.4$ & $81.4$\\ 
\midrule
{\em Our Models}\\
\textsc{IncSQL} (static oracle)  & $\mathbf{76.1}$ & $82.5$ & $\mathbf{75.5}$ & $81.6$ \\
\textsc{IncSQL} (non-det. oracle, ``order-matters'' only) & $75.4$ & $82.2$ & $75.1$ & $81.8$\\
\textsc{IncSQL} (non-det. oracle) & $49.9$ & $84.0$ & $49.9$ & $83.7$ \\
\textsc{IncSQL} (non-det. oracle) + EG (5) & $51.3$ & $\mathbf{87.2}$ & $51.1$ & $\mathbf{87.1}$ \\
\bottomrule
\end{tabular}
\caption{Dev and Test accuracy (\%) on WikiSQL. Acc$_\text{lf}$ refers to logical form accuracy and Acc$_\text{ex}$ refers to execution accuracy. ``+ EG (5) '' indicates execution-guided decoding with beam size of $5$. }
\label{result-table}
\end{table}

\begin{table}[t]
\centering
\small
\begin{tabular}{rcccc}
\toprule
Training Oracles  & w/o EG  & + EG (1) & + EG (3) & + EG (5) \\
\midrule
static            & $81.6$ & $83.5$ & $86.4$ & $86.7$ \\
non-determinstic  & $83.7$ & $86.0$ & $87.1$ & $87.1$ \\
\midrule
Speed (instances per second) & $48.3$ & $30.1$ & $8.2$ & $4.4$ \\
\bottomrule
\end{tabular}
\caption{Execution accuracy (\%) and decoding speed of our models on the test set of WikiSQL, with varying decoding beam size. %
The notation ``+ EG ($k$)'' is as in \Tabref{result-table}.
}
\label{result-table-eg}
\end{table}

We also report the performance of the static oracle model with execution-guided decoding in \Tabref{result-table-eg}.
It comes closely to the performance of the non-deterministic oracle model, but requires a larger beam size,
which translates to an increase in the decoding time.

\subsection{Results on other NL2SQL datasets}

To test whether our model can generalize to other datasets, we perform experiments with the ATIS dataset~\citep{Price1990, Dahl1994}. %
ATIS has more diverse SQL structures, including queries on multiple tables and nested queries.
To be compatible with our task setting, we only retain examples in the ATIS dataset that are free of nested queries, containing only \texttt{AND} operations and no \texttt{INNER JOIN} operators. 
We perform table joins and create a single table to be included in the input to our models along with the natural language question.
The reduced dataset consists of 933 examples, with 714/93/126 examples in the train/dev/test split, respectively.

Our models trained with the static and non-deterministic oracles (without \texttt{ANYCOL}) achieve accuracy of $67.5\%$ and $69.1\%$ on the test set, respectively. 
The improvement gained from using non-deterministic oracles during training validates our previous hypothesis: ATIS is a much smaller dataset compared with WikiSQL, therefore explicitly addressing ``order-matters'' helps here. We didn't apply \texttt{ANYCOL} due to the nature of ATIS data.

\begin{table}[t]
\centering
\small
\begin{tabular}{lcccc}
\toprule
\multirow{2}{*}{Model}& \multicolumn{2}{c}{Dev} & \multicolumn{2}{c}{Test} \\ \cmidrule{2-5}
  & Acc$_\text{lf}$    & Acc$_\text{ex}$ & Acc$_\text{lf}$    & Acc$_\text{ex}$\\
\midrule
\textsc{IncSQL} (static oracle)  & $87.1$ & $88.2$ & $65.9$ & $67.5$\\
\textsc{IncSQL} (non-det. oracle, ``order-matters'' only) & $88.1$ & $89.2$ & $68.3$ & $69.1$ \\
\bottomrule
\end{tabular}
\caption{Dev and Test accuracy (\%) of the models on the reduced ATIS dataset, where Acc$_\text{lf}$ refers to logical form accuracy and Acc$_\text{ex}$ refers to execution accuracy. }
\label{result-table-ATIS}
\end{table}

\section{Related Work}

WikiSQL, introduced by \citet{zhongSeq2SQL2017}, is the first large-scale dataset with annotated pairs of natural language queries and their corresponding SQL forms on a large selection of table schemas. While its coverage of SQL syntax is weaker than previous datasets such as ATIS~\citep{Price1990, Dahl1994} and GeoQuery~\citep{Zelle1996}, 
WikiSQL is highly diverse in its questions, table schemas and contents.
This makes it an attractive dataset for neural network modeling.
Indeed, a large number of recent works have already been evaluated on WikiSQL~\citep{chenglong,xu2017sqlnet, pshuang2018PT-MAML,yu2018naacl,2018executionguided,dong2018acl,McCann2018decaNLP}. 

NL2SQL is a special case of \emph{semantic parsing}. The task of semantic parsing maps natural language to a logical form representing its meaning, and has been studied extensively by the natural language processing community (see \citealt{liang2016executable} for a survey). The choice of meaning representation is usually task-dependent, including lambda calculus~\citep{wong2007acl}, lambda dependency-based compositional semantics \citep[$\lambda$-DCS]{Liang2013}, and SQL~\citep{zhongSeq2SQL2017}.
Neural semantic parsing, on the other hand, views semantic parsing as a sequence generation problem.
It adapts deep learning models such as those introduced by~\citet{Sutskever2014,Bahdanau2015,Vinyals2015}.
Combined with data augmentation~\citep{Jia2016, Iyer2017} or reinforcement learning~\citep{zhongSeq2SQL2017}, sequence-to-sequence with attention and copying has already achieved state-of-the-art results on many datasets including WikiSQL.

The meaning representation in semantic parsing usually has strict grammar syntax, as opposed to target sentences in machine translation.
Thus, models are often constrained to output syntactically valid results.
\citet{dong2016acl,dong2018acl} propose models that generate tree outputs through hierarchical decoding
and models that use sketches to guide decoding, but they do not explicitly deal with grammar constraints.
In contrast, \citet{yin17acl} and \citet{krishnamurthy2017neural} directly utilize grammar productions during decoding.

Training oracles have been extensively studied for the task of syntactic parsing, where incremental approaches are common \citep{goldberg13tacl}.
For syntactic parsing, due to the more structurally-constrained nature of the task and clearly-defined partial credits for evaluation, dynamic oracles allow the parsers to find optimal subsequent actions even if they are in some sub-optimal parsing states \citep{goldberg12coling,goldberg14tacl,cross16emnlp}. 
In comparison, non-deterministic oracles are defined for the optimal parsing states that have potential to reach a perfect terminal state. 
To the best of our knowledge, our work is the first to explore non-deterministic training oracles for incremental semantic parsing. 

\section{Conclusions}
\label{sec:conclusion}

In this paper, we introduce a sequence-to-action incremental parsing approach for the NL2SQL task. 
With the observation that multiple SQL queries can have the same or very similar semantics corresponding to a given natural language question, 
we propose to use non-deterministic oracles during training. 
On the WikiSQL dataset, our model trained with the non-deterministic oracles achieves an execution accuracy of $83.7\%$, which is $2.3\%$ higher than the current state of the art. 
We also discuss using execution-guided decoding in combination with our model. 
This leads to a further improvement of $3.4\%$, achieving a new state-of-the-art $87.1\%$ execution accuracy on the test set. 

To the best of our knowledge, our work is the first to use non-deterministic oracles for training incremental semantic parsers. 
Designing such non-deterministic oracles requires identification of multiple correct transition sequences for a given training instance,
and an algorithm that decides the possible continuations for any intermediate state that will lead to one of the desired terminal states. 
We have shown promising results for WikiSQL and filtered ATIS dataset and it would be interesting to extend our work to other more complex NL2SQL tasks and to other semantic parsing domains. %

\small
\bibliography{ref}
\bibliographystyle{iclr2019_conference}

\end{document}